\documentclass[conference]{IEEEtran}
\IEEEoverridecommandlockouts
\usepackage{cite}
\usepackage{amsmath,amssymb,amsfonts}
\usepackage{algorithmic}
\usepackage{graphicx}
\usepackage{textcomp}
\usepackage{xcolor}
\usepackage{multirow}
\usepackage{mmstyles}

\def\BibTeX{{\rm B\kern-.05em{\sc i\kern-.025em b}\kern-.08em
    T\kern-.1667em\lower.7ex\hbox{E}\kern-.125emX}}
\begin{document}

\title{TCSloT: Text Guided 3D Context and Slope Aware Triple Network for Dental Implant Position Prediction\\
}

\author{
\IEEEauthorblockN{1\textsuperscript{st} Xinquan Yang}
\IEEEauthorblockA{\textit{College of Computer Science and} \\
\textit{Software Engineering.}\\
\textit{Shenzhen University}\\
Shenzhen, China \\
xinquanyang99@gmail.com}
\and
\IEEEauthorblockN{2\textsuperscript{nd} Jinheng Xie}
\IEEEauthorblockA{\textit{Show Lab} \\
\textit{National University of Singapore}\\
Singapore \\
sierkinhane@gmail.com}
\and
\IEEEauthorblockN{3\textsuperscript{rd} Xuechen Li}
\IEEEauthorblockA{\textit{College of Computer Science and} \\
\textit{Software Engineering.}\\
\textit{Shenzhen University}\\
Shenzhen, China \\
timlee@szu.edu.cn}
\and
\IEEEauthorblockN{4\textsuperscript{th} Xuguang Li}
\IEEEauthorblockA{\textit{Department of Stomatology} \\
\textit{Shenzhen University General Hospital}\\
Shenzhen, China \\
lixuguang@szu.edu.cn}
\and
\IEEEauthorblockN{5\textsuperscript{th} Linlin Shen}
\IEEEauthorblockA{\textit{College of Computer Science and} \\
\textit{Software Engineering.}\\
\textit{Shenzhen University}\\
Shenzhen, China \\
llshen@szu.edu.cn}
\and
\IEEEauthorblockN{6\textsuperscript{th} Yongqiang Deng}
\IEEEauthorblockA{\textit{Department of Stomatology} \\
\textit{Shenzhen University General Hospital}\\
Shenzhen, China \\
qiangyongdeng@sina.com}
\thanks{(Corresponding author: Linlin Shen.)}
}
\maketitle

\begin{abstract}
In implant prosthesis treatment, the surgical guide of implant is used to ensure accurate implantation. However, such design heavily relies on the manual location of the implant position. When deep neural network has been proposed to assist the dentist in locating the implant position, most of them take a single slice as input, which do not fully explore 3D contextual information and ignoring the influence of implant slope. In this paper, we design a Text Guided 3D Context and Slope Aware Triple Network (TCSloT) which enables the perception of contextual information from multiple adjacent slices and awareness of variation of implant slopes. A Texture Variation Perception (TVP) module is correspondingly elaborated to process the multiple slices and capture the texture variation among slices and a Slope-Aware Loss (SAL) is proposed to dynamically assign varying weights for the regression head. Additionally, we design a conditional text guidance (CTG) module to integrate the text condition (i.e., left, middle and right) from the CLIP for assisting the implant position prediction. Extensive experiments on a dental implant dataset through five-fold cross-validation demonstrated that the proposed TCSloT achieves superior performance than existing methods.

\end{abstract}
% Dental Implant \and Deep Learning \and Text Guided Detection \and Cross-Modal Interaction

\section{Introduction}
Dental implant is a common surgical procedure in oral and maxillofacial surgery~\cite{varga2020guidance}, in which the surgical guide plays an important role in precise bone drilling and implant placement~\cite{gargallo2021intra}~\cite{vinci2020accuracy}. Generally, the design of the surgical guide relies on the patient’s panoramic radiographic image, or cone beam computed tomography (CBCT), image to estimate the implant position. However, traditional surgical guide design requires extensive manual work for preoperative planning, which significantly increases the cost of template fabrication~\cite{liu2021transfer}. Besides, the design of the surgical guide heavily depends on the subjective experiences of dentists, is thus quite inefficient and tedious. In contrast, Artificial intelligence (AI) based systems have demonstrated a strong ability to quickly locate the implant position, which potentially improves the efficiency of surgical guide design.

\begin{figure*}[]
\centering
\includegraphics[width=0.85\linewidth]{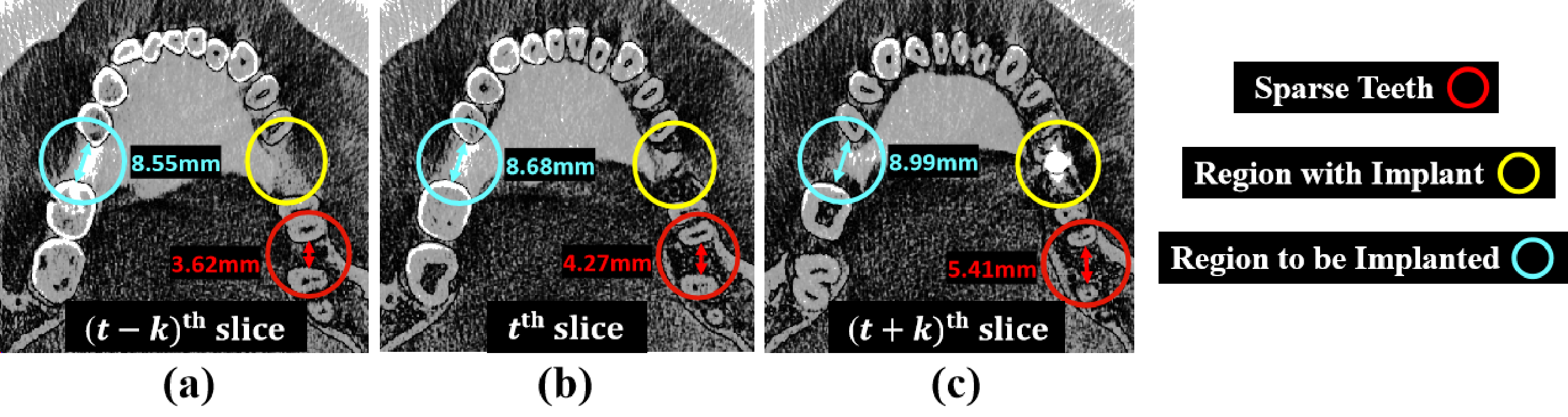}
\caption{Comparison of different 2D slices at tooth crown.} \label{fig_sparse_teeth}
\end{figure*}

\begin{figure}[]
\centering
\includegraphics[width=1.0\linewidth]{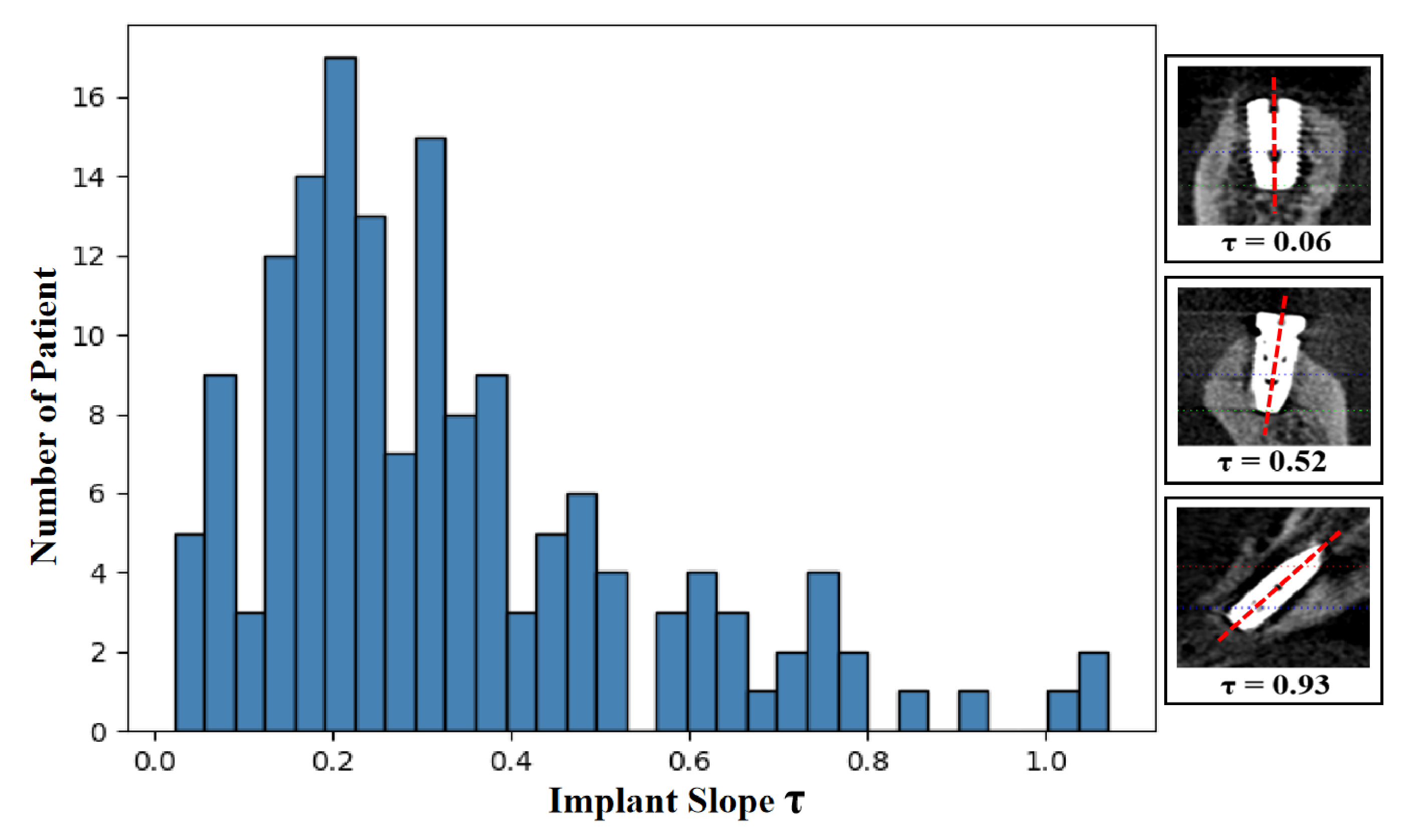}
\caption{The distribution of implant slope in the dental implant dataset.} \label{fig_slope}
\end{figure}

% Jinheng
Since 3D neural networks require high computation costs, recent works~\cite{kurt2021deep}\cite{widiasri2022dental} tend to directly predict implant position on a single 2D slice extracted from the 3D CBCT data. However, though such a manner is relatively efficient, they discard the 3D contextual information, e.g., the relationship among adjacent slices, potentially leading to worse implant precision. Consider only one 2D slice like Fig.~\ref{fig_sparse_teeth}(b), the texture of the sparse teeth region and the region to be implanted, share high similarity with the real implant region, which may result in the inaccurate prediction of implant position. When multiple slices like Figs.~\ref{fig_sparse_teeth}(a), (b), and (c) are involved, the distances of the neighboring teeth at the sparse teeth region change faster than the real implant region, and the texture of the implant will emerge in the lower slice. These contextual information can help to distinguish the sparse teeth and the region with implant, avoiding inaccurate detection. Besides, as the width of each patient's alveolar bone is different, the implant should be tilted with a suitable slope during operation (See Fig.~\ref{fig_slope}). The larger the slope of the implant, the greater the displacement of implant positions among involved 2D slices. Hence, contextual information among slices and implant slope are required for an accurate operation of dental implant.

% Recent literature works~\cite{}~\cite{} convert the 3D CBCT data into a series of 2D slices and finish the implant position prediction on them, avoiding the usage of 3D neural network which requires high computation costs. However, such a method do not consider the relationship between different slices, cause the loss of 3D context. As shown in Fig.~\ref{fig_sparse_teeth}, the texture of sparse teeth region and the region has been implanted are similar to the real implant region, where the prediction network will easily generate error detection. While switching to the view of multiple slices, the texture variation of the sparse teeth is faster than the real implant and the texture of completed implant will emerge. These contextual information can help the prediction model to distinguish the sparse teeth or completed implant region that avoids error detection. Moreover, in the clinical, the implant will be tilted at different angles since the available bone widths of patients are various (See Fig.~\ref{fig_slope}). It will cause different implant slopes. The larger the slope of the implant, the greater the displacement of implant position at the 2D slice. Therefore, a detection model that can perceive contextual information from different slices is needed.

To mitigate above problems, TSIPR~\cite{yang2023two} introduced a two-stream network that uses an implant region detector (IRD) to locate the most probable implant region to filter out the error detections generated by the regression network. ImplantFormer~\cite{yang2022implantformer} proposed to fit the centerline of implant using a series of prediction results of the 2D slices during inference, which remedies the 3D context loss by associating the results of independent 2D slices. TCEIP~\cite{yang2023tceip} introduced the direction embedding form CLIP~\cite{radford2021learning} to guide the prediction model where to implant, thus avoids the false alarms. However, both the contextual information among adjacent slices and implant slope are not studied in these works.

% TSIPR heavily relies on the accuracy of IRD which locates an implant region, but the IRD may locate the sparse teeth region in some 2D slices.

% As shown in Fig.~\ref{fig_network}, the texture among different slices includes rich context, e.g., the variation of the teeth region, which can greatly inspire the detection model to recognize the implant or sparse teeth region (The distance between the neighboring teeth in the sparse teeth region has a big variation than the implant region). Therefore, a detection model that can perceive multiple slices is needed.

% Dental-YOLO~\cite{} utilizes the 2D sagittal view of CBCT to measure the oral bone, e.g., the alveolar bone, and determine the implant position indirectly. TSIPR~\cite{} introduces a two-stream network that uses an implant region detector (IRD) to locate the most probable implant region to filter out the error detection generated by the regression network.
% To tackle this issue, ImplantFormer~\cite{} proposes to fit the centerline of implant using a series of prediction results of the 2D slices in inference. It remedies the 3D context loss by associating all the ind  ependent 2D slices and will not introduce large computation costs. Nevertheless, we argue that

% ImplantFormer~\cite{} predicts the implant position using the 2D axial view of tooth crown images and projects the prediction results back to the tooth root by the space transform algorithm.

In this paper, we design a Text Guided 3D Context and Slope Aware Triple Network (TCSloT), to integrate the perception of contextual information from multiple adjacent slices and awareness of variation of implant slopes. Specifically, we construct a triplet of $(t-k)^{th}$, $t^{th}$, and $(t+k)^{th}$ slices as input. A Texture Variation Perception (TVP) module is proposed to process the triplet and capture the texture variation among slices. Besides, a Slope-Aware Loss (SAL) is proposed to dynamically assign different weights for the regression head, which takes into account the influence of different slopes for the offset of implant position. The dynamic weights are calculated based on the implant slope. Additionally, we design a Conditional Text Guidance (CTG) module, which consists of a conditional text interaction and the knowledge alignment module~\cite{yang2023tceip}, to integrate the text condition from the CLIP for assisting the implant position prediction. The proposed CTG module enables the TCSloT to perform well in the case of multiple missing teeth.

% In this paper, we  to design a Slope-Aware Trident Networks (SA-TridentNet) which perceives the contextual information from different slices. Specifically, we construct triplets of the last, current, and next slice as network inputs. To better capture the texture variation from different slices, we design a Texture Variation Perception (TVP) module, which fuses the feature map of different slices. It consists of a self-attention block and a residual connection. The self-attention block pays attention to the texture variation across slices for forecasting while the residual connection provides the basic information and features of the current slices. A Slope Aware Loss (SAL) is proposed to dynamically assign varying weights for the regression head, which takes into account the influence of different slopes for the offset of implant position. The dynamic weights are calculated by the implant slope. Additionally, we follow TCEIP to integrate the text condition from the CLIP into our network, and lightweight the cross-modal interaction. It enables the SA-TridentNet to perform well in the case of multiple missing teeth.

% Moreover, we observe from the CBCT data that the implants in different patients' mouths have various postures (see Fig. 1), caused by the different implant angles. Therefore, each implant has different slopes and the slope of implant will influence the change rate of implant position at the 2D axial view. Motivated by this observation, we introduce

We conducted extensive experiments on a dental implant dataset~\cite{yang2022implantformer} and five-fold cross-validation was performed to validate our network performance. Main contributions of this paper are summarized as follows:
\begin{itemize}
\item To the best of our knowledge, the proposed TCSloT is the first multi-slices based implant position regression network that simultaneously perceives contextual information and implant slope.
\item A Texture Variation Perception (TVP) module and a Slope-Aware Loss (SAL) are devised to integrate the perception of contextual information from multiple adjacent slices and information of implant slopes, respectively.
% \item A Texture Variation Perception (TVP) module is devised to capture the texture variation across slices.
% \item A Slope Aware Loss (SAL) is designed to alleviate the influence of different slopes for the offset of implant position.
\item We design a Conditional Text Guidance (CTG) module to integrate the text condition into the TCSloT to assist the implant position prediction.
\item Extensive experiments show that the proposed TCSloT achieves superior performance than the previous methods.
\end{itemize}

\begin{figure*}
\centering
\includegraphics[width=1.0\linewidth]{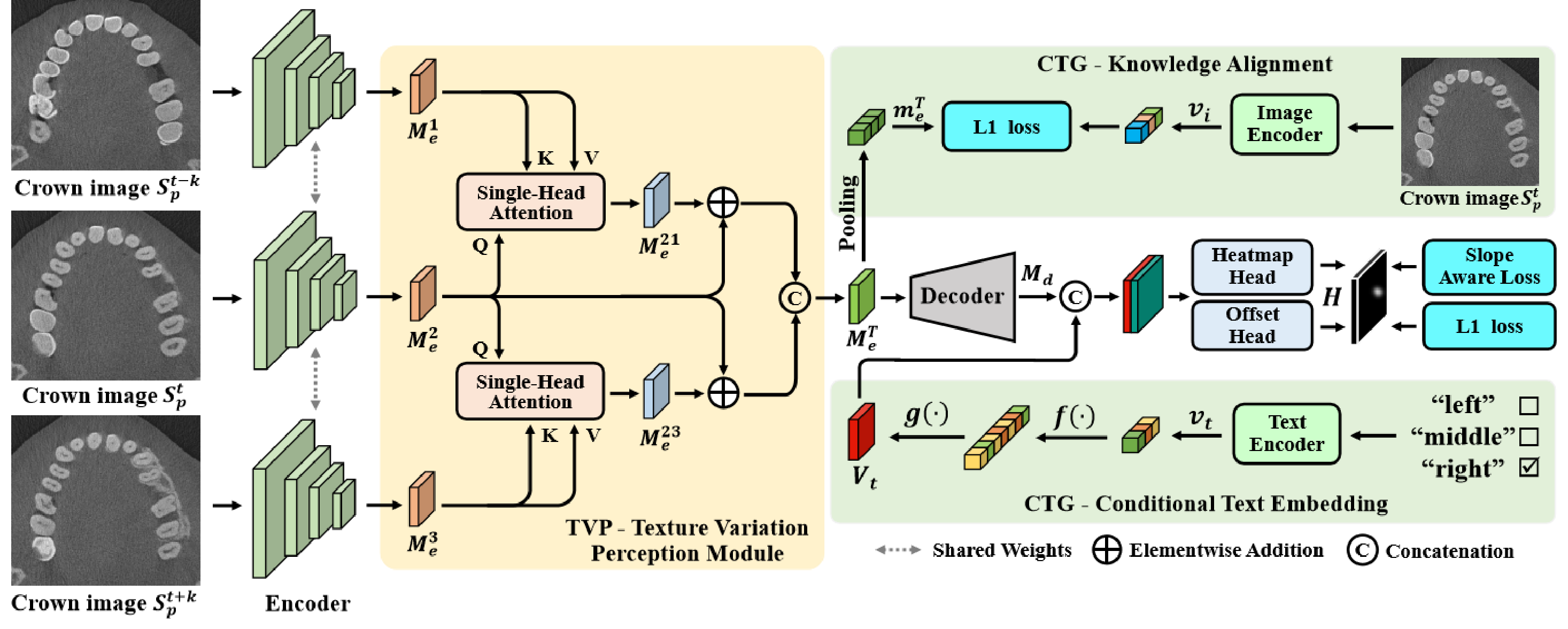}
\caption{ The overview of the proposed TCSloT, which consists of four parts, i.e., Encoder and Decoder, Texture Variation Perception Module (TVP), ConditionalText Guidance (CTG) Module, and Regression Head.} \label{fig_network}
\end{figure*}

\section{Methods}
Using tooth crown image to predict the implant position has shown its effectiveness~\cite{yang2022implantformer}. In this work, we follow this paradigm to train the proposed network. An overview of TCSloT is presented in Fig.~\ref{fig_network}, which mainly consists of four parts: i) Encoder and Decoder, ii) Texture Variation Perception Module (TVP), iii) Conditional Text Guidance (CTG) Module, and iv) Regression Head. After obtaining the predicted coordinates of the implant at the tooth crown, we project them from the tooth crown to tooth root area by the space transformation algorithm~\cite{yang2022implantformer} to obtain the actual implant position. Next, we will introduce these modules in detail.

% Three shared weights backbones extract the image features respectively. Then, the output features are input into the TVP module for information interaction. The ouput of TVP will be fead into the decoder to recover high-resolution features.

%  $\mathbb{R}^{\frac{H}{4}\times \frac{W}{4}\times \hat C}$
\subsection{Encoder and Decoder}
The encoder of TCSloT can be replaced by any available network. To make a fair comparison with other detectors, the widely used ResNet~\cite{he2016deep} is employed. Specifically, we use three ResNet-50 models with shared weights as the encoder, which enables the prediction network to perceive the texture variation from different slices. Given a 2D slice of tooth crown $S_p^t\in \mathbb{R}^{H\times W\times C}$ of patient $p$, we extract the upper $k$-th slice, the current slice and the lower $k$-th slice to form a triplet ($S_p^{t-k}$, $S_p^{t}$, $S_p^{t+k}$) as input. $k$ is set as 7 in this work, which is determined by the ablation experiment. A set of feature maps, i.e., $\{\mM_e^1, \mM_e^2, \mM_e^3\}$, can be accordingly extracted from the output of ResNet. Each feature map has a spatial and channel dimension $\mathbb{R}^{\frac{H}{4}\times \frac{W}{4}\times \hat C}$.

To ensure fine-grained heatmap regression, we design a decoder to recover high-resolution features. The decoder consists of three deconvolution layers. It consecutively upsamples the output feature map $\mM_e^T$ of TVP, as high-resolution feature representations, in which the recovered features $\mM_d$ can be extracted.
% Feature maps $M_3^d\in \mathbb{R}^{128\times 128\times C}$ will be further employed in the proposed modules.

% As shown in Fig.~\ref{fig_network}, the texture among different slices includes rich context, e.g., the variation of the teeth region, which can greatly inspire the detection model to recognize the implant or sparse teeth region (The distance between the neighboring teeth in the sparse teeth region has a big variation than the implant region). Therefore, a detection model that can perceive multiple slices is needed
\subsection{Texture Variation Perception Module (TVP)}
Given the output features of encoder for different slices, we consider two important characteristics to perceive the texture variations. One is the contextual information among slices, which can guide the detection model to discriminate the implant, completed implant or sparse teeth region. As shown in Fig.~\ref{fig_sparse_teeth}, when multiple slices are concerned, the distances between the neighboring teeth at the sparse teeth region change faster than the implant region, and the texture of the region with implant appear in the lower slice. The basic semantic information of the current slice is the other important characteristics to help the prediction model to locate the implant position.

We thus design a Texture Variation Perception (TVP) module to fuse the features of different slices using the cross-attention and residual connection. The architecture of TVP is given in Fig.~\ref{fig_network}. We implement the cross-attention module by two independent single-head self-attention modules, which takes the $t^{th}$ slice as the query and the upper$(t-k)^{th}$ or lower $(t+k)^{th}$ slice as the key and value:
\begin{equation}
\mM_e^{21}=softmax(\frac{\mM_e^2{\mM_e^1}^T}{\sqrt{d_{b1}}})\mM_e^1,
\end{equation}
\begin{equation}
\mM_e^{23}=softmax(\frac{\mM_e^2{\mM_e^3}^T}{\sqrt{d_{b2}}})\mM_e^3,
\end{equation}
where $\sqrt{d_{b1}}$ and $\sqrt{d_{b2}}$ are the dimension of $\mM_e^1$ and $\mM_e^3$, respectively. By this means, features of multiple slices can be fused to capture the tendency of texture variation. We add the original feature of the current slice to the outputs of the cross-attention module $\{\mM_e^{21}, \mM_e^{23}\}$  by the residual connection. It will provide the basic information for the fusion feature to improve the predicting robustness. In the end, we concatenate these two fusion features to generate the output $\mM_e^T$:
\begin{equation}
\mM_e^T=\textbf{concat}[\mM_e^{21}+\mM_e^2, \mM_e^{23}+\mM_e^2].
\end{equation}

\subsection{Conditional Text Guidance Module (CTG)}
Clinically, the patient may have multiple missing teeth to be implanted. When there is missing teeth on both sides of alveolar bone, the doctor usually choose one side for treatment first, such that the patient can use the other side for chewing. Moreover, the width of the anterior alveolar bone is much narrower than that of the bilateral alveolar bone, which needs to be considered separately when selecting implants. Therefore, we divide teeth regions into three categories, i.e., left, middle, and right and use target region as a prior information to help the network decide the implant position. While such information is not available to the network, it can be integrated by inputting the text provided by dentists, when using the network for implant position regression. To this end, we design a conditional text guidance (CTG) module, which consists of a conditional text interaction and the knowledge alignment module.

% In this work, we follow TCEIP to integrate the text condition from the CLIP~\cite{} into our network. Instead of using the cross-modal attention module to finish the cross-modal interaction in both large resolution feature map of backbone and decoder, we directly combine the text embedding with the last layer of decoder, which considering the shallow layer of backbone do not contains rich semantic information and removing the interaction module only with a slight performance loss.

We implement conditional text interaction by two steps: 1) The directional vocabulary, i.e., 'left', 'middle', or 'right', is processed by the CLIP Text Encoder to obtain a conditional text embedding $\vv_t \in \mathbb{R}^{1\times D}$. To interact with the image features from the encoder, a series of transformation $f(\cdot)$ and $g(\cdot)$ are applied to $v_t$ as follow:
\begin{equation}
\mV_t=g(f(\vv_t)) \in \mathbb{R}^{H\times W\times 1},
\end{equation}
where $f(\cdot)$ repeats text embedding $v_t$ from $\mathbb{R}^{1\times D}$ to $\mathbb{R}^{1\times HW}$ and $g(\cdot)$ reshapes it to $\mathbb{R}^{H\times W\times 1}$. This operation ensures better interaction between image and text in the same feature space. 2) $\mV_t$ is concatenated with the last layer of decoder to fuse both modal features. By this means, TCSloT is able to generate direct prediction according to the conditional text.

As there is a knowledge gap between the text embedding extracted by the CLIP and image features initialized by ImageNet pre-training, we follow~\cite{yang2023tceip} to perform the knowledge alignment, which gradually aligns image features to the feature space of pre-trained CLIP. The knowledge alignment is formulated as follows:
\begin{equation}
    \mathcal{L}_{align} = |\vm_e^T - \vv_i|,
\end{equation}
where $\vm_e^T\in \mathbb{R}^{1\times D}$ is the output feature map of TVP after the attention pooling operation~\cite{dosovitskiy2020image} and dimension reduction with convolution. $\vv_i\in \mathbb{R}^{1\times D}$ is the image embedding extracted by the CLIP Image Encoder. Using this criteria, the encoder of TCSloT approximates the CLIP image encoder and consequently aligns the image features of the encoder with the CLIP text embeddings.

\subsection{Slope-Aware Loss (SAL)}
% In the clinical, the width of alveolar bone in different patients are different, which resul
% We observe an important characteristic from the CBCT data that the implants have various postures in patients' mouths (see Fig.~\ref{fig_slope}), caused by the different implant angles.

% It can be observed from the CBCT data that the implant slopes among patients are significantly different. Therefore, each implant has different slopes and it will influence the change rate of implant position at the 2D axial view, i.e., the larger the slope, the greater the displacement of the implant position. We visualize the implant slope of our dental implant dataset in Fig.~\ref{fig_slope}, from which we can observe that the slope of most implants are relatively flat, while a few are significantly large. This phenomenon is consistent with clinical practice that most patients have sufficiently wide alveolar bone for dental implant, without the need for tilting the implant.

As the width of each patient's alveolar bone is different, the implant should be tilted with a suitable slope during operation, which will influence the variation of implant position at the 2D axial view, i.e., the larger the slope, the greater the displacement of the implant position. We use the space transformation algorithm~\cite{yang2022implantformer} to calculate the implant slope:
\begin{equation}
s_1=\frac{N^p_c \sum_{i=1}^{N^p_c} x_i z_i-\sum_{i=1}^{N^p_c} x_i \times \sum_{i=1}^{N^p_c} z_i}{N^p_c \sum_{i=1}^{N^p_c} z_i^2-\sum_{i=1}^{N^p_c} z_i \times \sum_{i=1}^{N^p_c} z_i},
\end{equation}
\begin{equation}
s_2=\frac{N^p_c \sum_{i=1}^{N^p_c} y_i z_i-\sum_{i=1}^{N^p_c} y_i \times \sum_{i=1}^{N^p_c} z_i}{N^p_c \sum_{i=1}^{N^p_c} z_i^2-\sum_{i=1}^{N^p_c} z_i \times \sum_{i=1}^{N^p_c} z_i},
\end{equation}
Here $s_1$, $s_2$ is the slope of the 3D spatial line. $(x_i, y_i, z_i),i\in [1,N^p_c]$ is the coordinate of implant position. $N^p_c$ is the total number of 2D slices. We simply add this two slopes together as the final slope $\tau$:
\begin{equation}
\tau = |s_1| + |s_2|.
\end{equation}

The distribution of implant slope in the dental implant dataset is shown in Fig.~\ref{fig_slope}, from which we can observe that the slope of most implants are small ($\tau<0.4$), while a few are significantly large. This phenomenon is consistent with clinical practice that most patients have sufficiently wide alveolar bone for dental implant, without the need for tilting the implant. Motivated by this observation, we introduce a Slope Aware Loss (SAL) to adopt adaptive weight for each implant, according to its slope. Specifically, we enforces the prediction model to pay more attention to the implant with large slope, as these implant positions deviate significantly across slices.
% It is more difficult to predict the implant position.

% Considering that simply applying $\tau$ as the weight to the loss of each implant will change the magnitude of the total losses. This
As the magnitude of $\tau$ is different with that of the total losses, which may disturb the balance between the loss of positive and negative samples and decrease the detection performance~\cite{yang2022real}. We follow~\cite{vsaric2019single}~\cite{saric2020warp} to normalize $\tau$ to $\hat \tau$ to keep the sum of total loss unchanged:
\begin{equation}
\hat \tau = \tau\cdot\frac{\sum_{i=1}^{N^p_c} \mathcal{L}^i_{h}}{\sum_{i=1}^{N^p_c} \tau\mathcal{L}^i_{h}},
\end{equation}
where $\mathcal{L}^i_{h}$ denotes the regression loss of implant position $i$, implemented by the Focal loss~\cite{lin2017focal}. We re-weight the regression loss of each object with $\hat \tau$ and all the 2D axial views slices belonging to the same implant share the same $\hat \tau$:
\begin{equation}
\mathcal{L}_{SAL}=\sum_{i=1}^{N^p_c} \hat \tau \mathcal{L}^i_{h}.
\end{equation}

\subsection{Regression Head}
The regression head of TCSloT consists of a heatmap head and a local offset head, which is used for predicting the implant position. The heatmap head generates an Gaussian heatmap $\mF\in[0,1]^{\frac{W}{g} \times \frac{H}{g}}$, where $g$ is the down sampling factor of the prediction and set as 4. Following the standard practice in CenterNet~\cite{zhou2019objects}, the ground-truth position is transformed into the heatmap $F$ using a 2D Gaussian kernel:
\begin{equation}
\mF_{xy}=\exp(-\frac{(x-{\tilde{t}_x})^2+(y-{\tilde{t}_y})^2}{2\sigma^2}),
\end{equation}
where $(\tilde{t}_x,\tilde{t}_y)$ is ground-truth annotation in $F$. $\sigma$ is an object size-adaptive standard deviation. The heatmap head is optimized by the proposed SAL loss.

The local offset head computes the discretization error caused by $g$, which is used to further refine the predicted location. The loss of the local offset $\mathcal{L}_{offset}$ is optimized by L1 loss~\cite{girshick2014rich}. The overall training loss of SA-TrdientNet is:
\begin{equation}
\mathcal{L}_{total}=\mathcal{L}_{SAL}+\mathcal{L}_{offset}+\mathcal{L}_{align}
\end{equation}

\section{Experiment}
\subsection{Dataset}
The dental implant dataset used for evaluation is collected from the Shenzhen University General Hospital (SZUH), and all the implant positions were annotated by three experienced dentists. Specifically, the dataset contains 154 patients, from which 3045 2D slices of tooth crown are selected. All the CBCT data were captured using the KaVo 3D eXami machine, manufactured by Imagine Sciences International LLC. Dentists firstly designed the virtual implant based on the CBCT data using the surgical guide design software. Then the implant position can be determined as the center of the virtual implant.

\subsection{Implementation Details}
Pytorch is used for model training and testing. We use a batch size of 8, Adam optimizer and a learning rate of 0.0005 for network training. Three data augmentation methods, i.e. random crop, random scale and random flip are employed. The network is trained for 80 epochs and the learning rate is divided by 10 at 40th and 60th epochs, respectively. All the models are trained and tested on the platform of TESLA A100 GPU. For the training of other baseline detectors, MMDetection library and ImageNet pre-training models are used.

\begin{figure}
\centering
\includegraphics[width=1.0\linewidth]{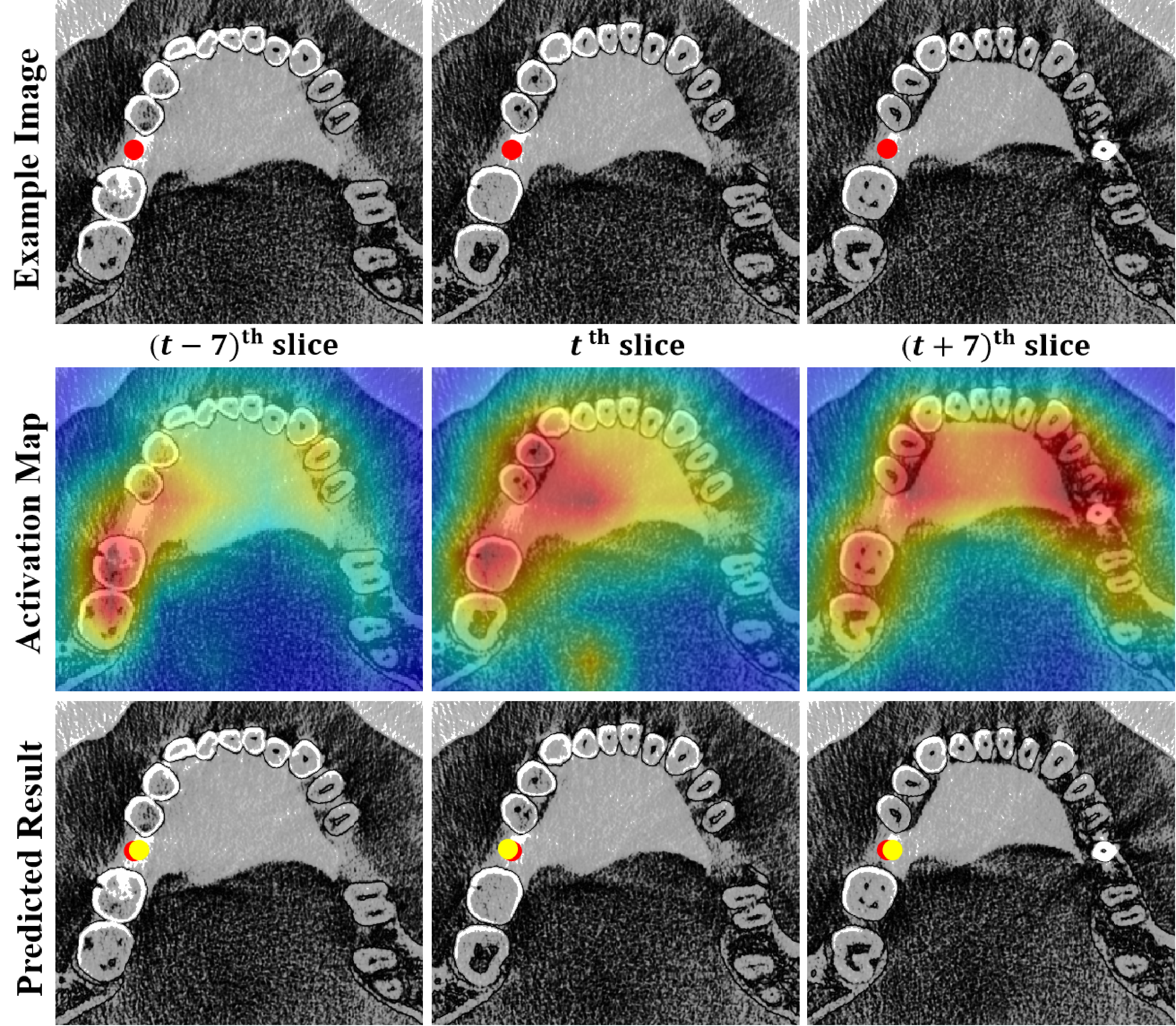}
\caption{Visualization of the attention map and prediction result at different slices. The yellow and red circles represent the predicted implant position and ground-truth position, respectively.} \label{fig_TVP}
\end{figure}

\begin{figure}
\centering
\includegraphics[width=0.95\linewidth]{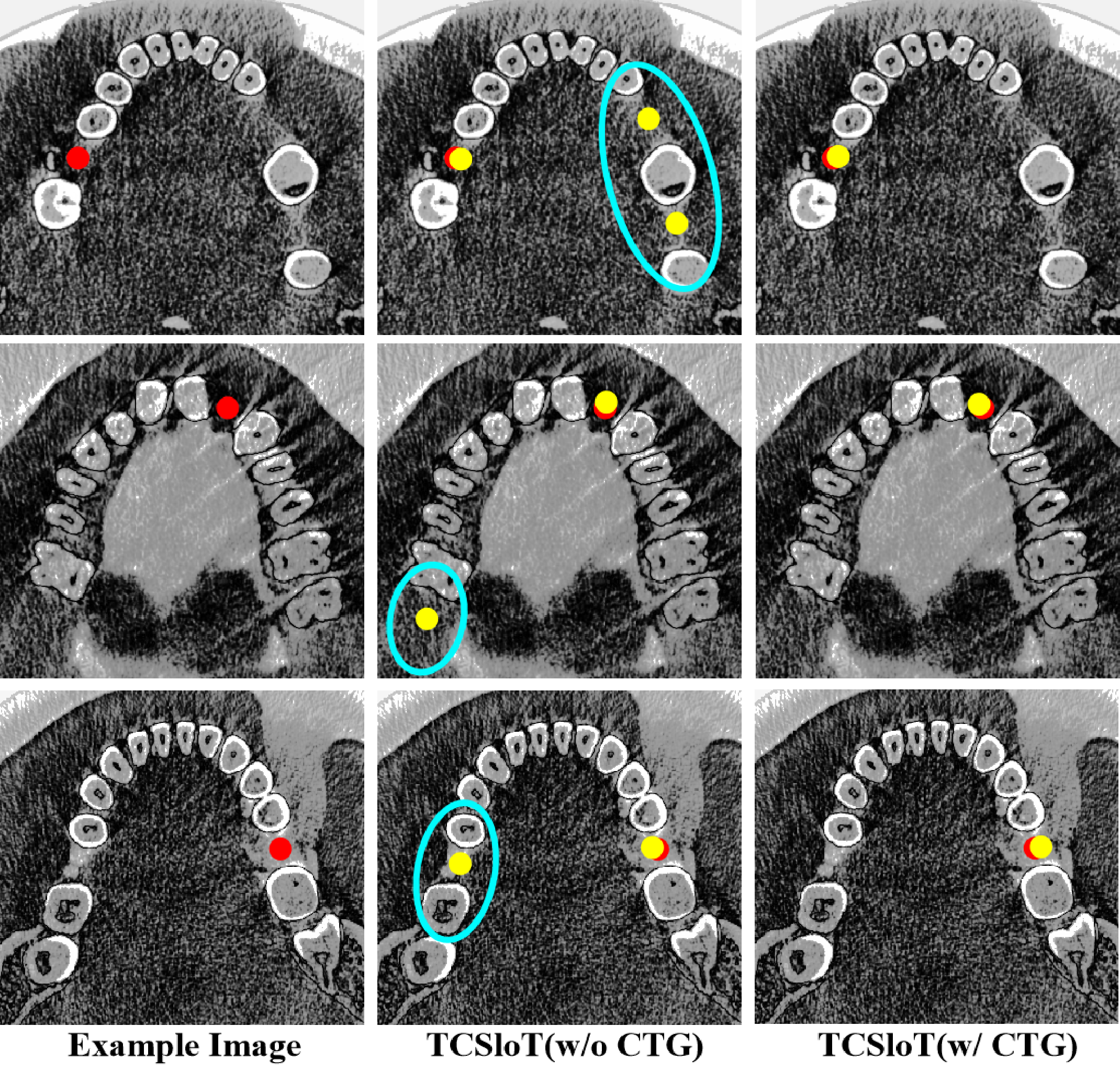}
\caption{Visual comparison of the TCSloT with or without the CTG module. The yellow and red circles represent the predicted implant position and ground-truth position, respectively. The blue ellipses denote false positive detections. From top to bottom, the text condition are left, middle, and right respectively.} \label{fig_CTG}
\end{figure}

\subsection{Evaluation Criteria}
The diameter of the implant is 3.5$\sim$5mm, and clinically the mean error between the predicted and ideal implant position is required to be less than 1mm, i.e., around 25\% of the size of implant. Therefore, $AP_{75}$ is used as the evaluation criteria. As the average radius of implants is around 20 pixels, a bounding-box with size $21\times21$ centered at the keypoint is generated. The calculation of AP is defined as follows:
\begin{equation}
Precition=\frac{TP}{TP+FP}
\end{equation}
\begin{equation}
Recall=\frac{TP}{TP+FN}
\end{equation}
\begin{equation}
AP=\int^1_0P(r)dr
\end{equation}
Here TP, FP and FN are the number of correct, false and missed predictions, respectively. P(r) is the PR Curve where the recall and precision act as abscissa and ordinate, respectively.

\subsection{Performance Analysis}
% \subsubsection{Ablation of Different Sampling Intervals}
\textbf{Sensitivity of Different Sampling Intervals.}
Using different sampling intervals $k$ will influence the extraction of contextual information in TCSloT. To determine a suitable value of $k$, we conduct a sensitivity analysis to compare the performance of setting different $k$. Results are listed in Table~\ref{table_samoling}. From the table we can observe that with the increment of $k$, the network performance increases gradually. When the sampling interval reaches seven, the network performance begins to decline. The reason for this phenomenon is that the large sampling interval ($k>7$) will cause the drastic variation of teeth texture between slices, which influences the prediction of implant position. Therefore, we set $k=7$ in this work.

\begin{table}[]
\caption{The ablation experiments of different sampling intervals in TCSloT.}\label{table_samoling}
\centering
\begin{tabular}{|llllllllll|l|}
\hline
\multicolumn{10}{|c|}{Sampling Intervals $k$} & \multirow{2}{*}{$AP_{75}$\%} \\ \cline{1-10}
\multicolumn{1}{|l|}{1} & \multicolumn{1}{l|}{2} & \multicolumn{1}{l|}{3} & \multicolumn{1}{l|}{4} & \multicolumn{1}{l|}{5} & \multicolumn{1}{l|}{6} & \multicolumn{1}{l|}{7} & \multicolumn{1}{l|}{8} & \multicolumn{1}{l|}{9} & 10 &                     \\ \hline
\multicolumn{1}{|l|}{\checkmark}  & \multicolumn{1}{l|}{}  & \multicolumn{1}{l|}{}  & \multicolumn{1}{l|}{}  & \multicolumn{1}{l|}{}  & \multicolumn{1}{l|}{}  & \multicolumn{1}{l|}{}  & \multicolumn{1}{l|}{}  & \multicolumn{1}{l|}{}  &    & 18.3$\pm$0.4      \\ \hline
\multicolumn{1}{|l|}{}  & \multicolumn{1}{l|}{\checkmark}  & \multicolumn{1}{l|}{}  & \multicolumn{1}{l|}{}  & \multicolumn{1}{l|}{}  & \multicolumn{1}{l|}{}  & \multicolumn{1}{l|}{}  & \multicolumn{1}{l|}{}  & \multicolumn{1}{l|}{}  &    & 17.9$\pm$0.6                     \\ \hline
\multicolumn{1}{|l|}{}  & \multicolumn{1}{l|}{}  & \multicolumn{1}{l|}{\checkmark}  & \multicolumn{1}{l|}{}  & \multicolumn{1}{l|}{}  & \multicolumn{1}{l|}{}  & \multicolumn{1}{l|}{}  & \multicolumn{1}{l|}{}  & \multicolumn{1}{l|}{}  &    & 19.1$\pm$0.3                     \\ \hline
\multicolumn{1}{|l|}{}  & \multicolumn{1}{l|}{}  & \multicolumn{1}{l|}{}  & \multicolumn{1}{l|}{\checkmark}  & \multicolumn{1}{l|}{}  & \multicolumn{1}{l|}{}  & \multicolumn{1}{l|}{}  & \multicolumn{1}{l|}{}  & \multicolumn{1}{l|}{}  &    & 18.8$\pm$0.3                     \\ \hline
\multicolumn{1}{|l|}{}  & \multicolumn{1}{l|}{}  & \multicolumn{1}{l|}{}  & \multicolumn{1}{l|}{}  & \multicolumn{1}{l|}{\checkmark}  & \multicolumn{1}{l|}{}  & \multicolumn{1}{l|}{}  & \multicolumn{1}{l|}{}  & \multicolumn{1}{l|}{}  &    & 19.5$\pm$0.4                     \\ \hline
\multicolumn{1}{|l|}{}  & \multicolumn{1}{l|}{}  & \multicolumn{1}{l|}{}  & \multicolumn{1}{l|}{}  & \multicolumn{1}{l|}{}  & \multicolumn{1}{l|}{\checkmark}  & \multicolumn{1}{l|}{}  & \multicolumn{1}{l|}{}  & \multicolumn{1}{l|}{}  &    & 19.7$\pm$0.2                     \\ \hline
\multicolumn{1}{|l|}{}  & \multicolumn{1}{l|}{}  & \multicolumn{1}{l|}{}  & \multicolumn{1}{l|}{}  & \multicolumn{1}{l|}{}  & \multicolumn{1}{l|}{}  & \multicolumn{1}{l|}{\checkmark}  & \multicolumn{1}{l|}{}  & \multicolumn{1}{l|}{}  &    & \textbf{20.4}$\pm$\textbf{0.3}                     \\ \hline
\multicolumn{1}{|l|}{}  & \multicolumn{1}{l|}{}  & \multicolumn{1}{l|}{}  & \multicolumn{1}{l|}{}  & \multicolumn{1}{l|}{}  & \multicolumn{1}{l|}{}  & \multicolumn{1}{l|}{}  & \multicolumn{1}{l|}{\checkmark}  & \multicolumn{1}{l|}{}  &    & 19.6$\pm$0.5                     \\ \hline
\multicolumn{1}{|l|}{}  & \multicolumn{1}{l|}{}  & \multicolumn{1}{l|}{}  & \multicolumn{1}{l|}{}  & \multicolumn{1}{l|}{}  & \multicolumn{1}{l|}{}  & \multicolumn{1}{l|}{}  & \multicolumn{1}{l|}{}  & \multicolumn{1}{l|}{\checkmark}  &    & 19.3$\pm$0.4                    \\ \hline
\multicolumn{1}{|l|}{}  & \multicolumn{1}{l|}{}  & \multicolumn{1}{l|}{}  & \multicolumn{1}{l|}{}  & \multicolumn{1}{l|}{}  & \multicolumn{1}{l|}{}  & \multicolumn{1}{l|}{}  & \multicolumn{1}{l|}{}  & \multicolumn{1}{l|}{}  & \checkmark   & 18.7$\pm$0.2                    \\ \hline
\end{tabular}
\end{table}

% \subsubsection{Component Ablation}
\textbf{Component Ablation.}
To demonstrate the effectiveness of the proposed modules, i.e., TVP, CTG, and SAL, we conduct ablation experiments for them. When the TVP module is removed, the output of the shared weights backbone is concatenated together and performs a channel reduction via a convolution. When the SAL loss is removed, focal loss is applied for the supervision of heatmap regression. The experimental results are listed in Table~\ref{table_ablation}. From the table we can observe that the proposed components are beneficial for TCSloT, among which the TVP module, the CTG module, and the SAL loss improve the performance by 3.9\%, 3.6\% and 3.8\%, respectively. When combining the TVP or CTG module with the SAL loss, the improvement of performance reaches 6.5\% and 7.0\%, respectively. When combining all these components, the improvement reaches 9.5\%.

\begin{figure}
\centering
\includegraphics[width=1.0\linewidth]{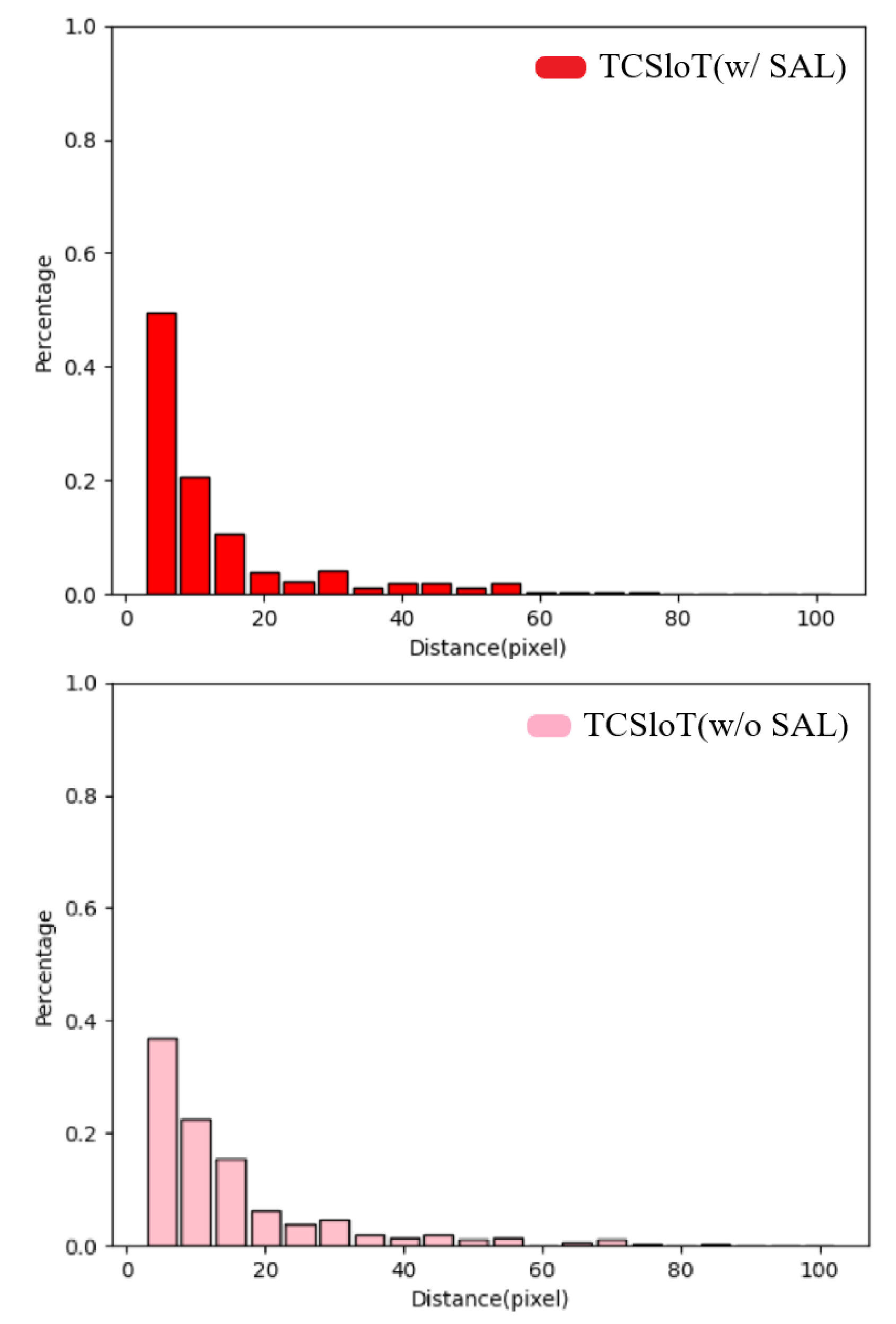}
\caption{The distribution of euclidean distances between the ground truth position and the predictions of TCSloT with or without the SAL loss.} \label{fig_SAL}
\end{figure}

In Fig.~\ref{fig_TVP}, we visualise the attention map and prediction result of TCSloT on a patient who has both implant and missing teeth, to further verify the effectiveness of the TVP module. We remove the CTG module to ensure no additional information is available. From the figure we can observe that the TVP module simultaneously perceives the implant and missing teeth region in the third slice, and only perceives the missing teeth region in the other slices. In addition, the prediction results of TCSloT without the CTG module do not generate false alarms in the first two slices. These visual results indicate that the TVP module enables the SA-TridentNte to include all the information of multiple slices.

In Fig.~\ref{fig_CTG}, we visualize the detection results of the TCSloT with or without the CTG module. The visual results indicates that the CTG module can greatly help the TCSloT to generate accurate implant position on the patient with multiple missing teeth. These visual results demonstrated the effectiveness of using text condition to assist the implant position prediction.

In Fig.~\ref{fig_SAL}, we visualise the euclidean distance between the ground truth position and the predictions of TCSloT with or without the SAL loss, to further demonstrate the effectiveness of the proposed SAL loss. The euclidean distances are summed in an interval of five. Smaller the distance, more accurate the implant position prediction. From the figure we can observe that the distance of both networks mainly distribute in the range of 0 to 15 pixels. For the TCSloT with the SAL loss, more than 70\% predictions are located within 10 pixels from groundtruth position, while for TCSloT without the SAL loss, only about 60\% predictions are located within 10 pixels. Considering that the diameter of implant is around 20 pixels, the predictions with distance more than 10 pixels are meaningless. Therefore, the SAL loss are greatly helpful for the prediction of implant position.

\begin{table}[]
\caption{The ablation experiments of the proposed components in TCSloT.}\label{table_ablation}
\centering
\begin{tabular}{|c|c|c|c|c|}
\hline
Network                        & TVP         & CTG          & SAL           & $AP_{75}$\%    \\ \hline
\multirow{8}{*}{TCSloT} &             &              &               & 10.9$\pm$0.2   \\ \cline{2-5}
                               & \checkmark  &              &               & 14.8$\pm$0.6   \\ \cline{2-5}
                               &             & \checkmark   &               & 14.5$\pm$0.3   \\ \cline{2-5}
                               &             &              & \checkmark    & 14.7$\pm$0.3   \\ \cline{2-5}
                               & \checkmark  &              & \checkmark    & 17.4$\pm$0.1   \\ \cline{2-5}
                               &             & \checkmark   & \checkmark    & 17.9$\pm$0.2   \\ \cline{2-5}
                               & \checkmark  & \checkmark   &               & 18.3$\pm$0.4   \\ \cline{2-5}
                               & \checkmark   & \checkmark   & \checkmark  &  \textbf{20.4}$\pm$\textbf{0.3} \\ \hline
\end{tabular}
\end{table}

\begin{table}[]
\caption{Comparison with the Text-guided Detection Methods.}\label{table_textmethod}
\centering
\begin{tabular}{|c|l|l|l|}
\hline
Network              & Encoder                    & $AP_{75}\%$                              & F1 Score \\ \hline
TransVG              & \multirow{4}{*}{ResNet-50} & 13.2$\pm$0.8                    & 13.1$\pm$0.6     \\ \cline{1-1} \cline{3-4}
VLTVG                &                            & 14.1$\pm$0.7                    & 13.9$\pm$0.7     \\ \cline{1-1} \cline{3-4}
JointNLT             &                            & 15.3$\pm$0.6                    & 15.1$\pm$0.6     \\ \cline{1-1} \cline{3-4}
SA-TrindentNet(ours) &                            & \textbf{20.4}$\pm$\textbf{0.3}  & \textbf{20.1}$\pm$\textbf{0.3}     \\ \hline
\end{tabular}
\end{table}

\begin{table*}[]
\caption{Comparison with other mainstream detectors.}\label{table_detector}
\centering
\begin{tabular}{|c|c|c|c|c|}
\hline
Method                             & Network                              & Encoder                    & $AP_{75}\%$ & F1 Score \\ \hline
\multirow{9}{*}{CNN-based}         & CenterNet                            & \multirow{7}{*}{ResNet-50} & 10.9$\pm$0.2   & 10.8$\pm$0.1          \\ \cline{2-2} \cline{4-5}
                                   & ATSS                                 &                            & 12.1$\pm$0.2   & 11.9$\pm$0.2          \\ \cline{2-2} \cline{4-5}
                                   & VFNet                                &                            & 11.8$\pm$0.8   & 11.8$\pm$0.1          \\ \cline{2-2} \cline{4-5}
                                   & RepPoints                            &                            & 11.2$\pm$0.1   & 11.1$\pm$0.3          \\ \cline{2-2} \cline{4-5}
                                   & ImplantFormer                        &                            & 11.5$\pm$0.3   & 11.3$\pm$0.3          \\ \cline{2-2} \cline{4-5}
                                   & TCEIP                                &                            & 17.8$\pm$0.3   & 17.6$\pm$0.2          \\ \cline{2-2} \cline{4-5}
                                   & TCSloT(ours)                  &                            & 20.4$\pm$0.3   & 20.1$\pm$0.3          \\ \cline{2-5}
                                   & MSPENet                              & \multirow{2}{*}{-}         & 15.4$\pm$0.3   & 15.2$\pm$0.2          \\ \cline{2-2} \cline{4-5}
                                   & TSIPR                                &                            & 15.7$\pm$0.4   & 15.6$\pm$0.3          \\ \hline
\multirow{6}{*}{Transformer-based} & Conditional DETR                     & \multirow{2}{*}{ResNet-50} & 12.7$\pm$0.2   & 12.6$\pm$0.2          \\ \cline{2-2} \cline{4-5}
                                   & Deformable DETR                      &                            & 12.8$\pm$0.1   & 12.5$\pm$0.1          \\ \cline{2-5}
                                   & ImplantFormer                        & ViT-B-ResNet-50            & 13.7$\pm$0.2   & 13.6$\pm$0.2          \\ \cline{2-5}
                                   & \multirow{3}{*}{TCSloT(ours)} & DeiT-B                     & 20.9$\pm$0.2   & 20.7$\pm$0.3          \\ \cline{3-5}
                                   &                                      & Swin-T                     & 21.6$\pm$0.2   & 21.3$\pm$0.4          \\ \cline{3-5}
                                   &                                      & SegFormer-B3               & \textbf{21.9}$\pm$\textbf{0.5}   & \textbf{21.8}$\pm$\textbf{0.2}          \\ \hline
\end{tabular}
\end{table*}

\textbf{Comparison with Text-guided Detection Methods.}
To further validate the performance of the proposed CTG module, we also compare TCSloT with other text-guided detction methods, e.g., TransVG, VLTVG and JointNLT. For a fair comparison, the widerly used ResNet-50 is employed as the encoder. Results are listed in Table~\ref{table_textmethod}. From the table we can observe that their $AP_{75}\%$ values are 13.2\%, 14.1\%, and 15.3\%, respectively, which are significantly lower than that of our proposed TCSloT (20.4\%). In terms of F1 score, TCSloT also performs the best. These results demonstrated the effectiveness of the proposed CTG module that uses text condition to assist the implant position prediction.

\textbf{Comparison with the Mainstream Detectors.}
To further demonstrate the superiority of our method, we compare the location performance of the proposed TCSloT with other detectors. As little useful texture is available around the center of implant, the anchor-based detectors cannot regress implant position successfully. Only the CNN-based anchor-free detectors (VFNet~\cite{zhang2021varifocalnet}, ATSS~\cite{zhang2020bridging}, RepPoints~\cite{yang2019reppoints}, CenterNet~\cite{zhou2019objects}), TCEIP~\cite{yang2023tceip}, MSPENet~\cite{yang2023two}, TSIPR~\cite{yang2023two}), transformer-based detectors (Conditional DETR~\cite{meng2021conditional}, Deformable DETR~\cite{zhu2020deformable}, ImplantFormer~\cite{yang2022implantformer}) are employed for comparison. For a fair comparison, ResNet-50 is employed as the encoder. Results are listed in Table~\ref{table_detector}.

From the table we can observe that text condition can significantly help improve the performance of implant position prediction, e.g., TCEIP achieved 17.8\% AP, which is 2.1\% higher than the best performed CNN-based network - TSIPR.  The proposed TCSloT achieves the best AP value - 20.4\%, among all benchmarks, which surpasses the TCEIP with a large gap. In addition to ResNet50, we have adopted other advanced encoder (e.g., DeiT~\cite{touvron2021training}, Swin Transformer~\cite{liu2021swin}, and SegFormer~\cite{xie2021segformer}), and the AP75 value are 20.9\%, 21.6\% and 21.9\%, respectively. These experimental results demonstrate the effectiveness of TCSloT.

% \begin{table}[]
% \caption{Comparison of advanced encoder.}\label{table_textmethod}
% \centering
% \begin{tabular}{|l|l|l|}
% \hline
% Methods                        & Encoder & $AP_{75}\%$ \\ \hline
% \multirow{3}{*}{SA-TridentNet} & DeiT-T    &    \\ \cline{2-3}
%                                & Swin-T    &    \\ \cline{2-3}
%                                & SegFormer     &    \\ \hline
% \end{tabular}
% \end{table}

% \textbf{Advanced Encoder.}
% To demonstrate the superiority of our method, we compare the location performance of the proposed SA-TridentNet with other detectors.

% \subsubsection{Comparison of Slice Views of Implant}
% \textbf{Comparison of Slice Views of Implant.}
% To validate the effectiveness of the proposed ImplantFormer, we compare four slice views of the actual implant with the predicted implant in Fig.~\ref{}. The slice view is the different longitudinal views of CBCT data of the implant, which can well demonstrate the direction and placement of the implant. For the predicted implant, a cylinder with a radius of 10 pixels centered at $Pos_r$, is generated, and the implant depth is manually set. The pixel value of CBCT data inside the cylinder area is set as 3100. For fair comparison, the longitudinal views from the CBCT data of both ground truth and predicted implant are selected at the same position.

% From the figure, we can observe that the implant position and direction generated by ImplantFormer are consistent with the ground truth implant, which confirms the accuracy of the proposed ImplantFormer.

\section{Conclusion}
In this work, we develop a ext Guided 3D Context and Slope Aware Triple Network (TCSloT) for CBCT data based implant position prediction, which takes a triple of the upper, current and lower slices as input. We use three backbones with shared weights to extract feature maps of different slices and design a Texture Variation Perception (TVP) module to fuse them. A conditional text guidance (CTG) module is designed to integrate the text condition into the TCSloT to assist the implant position prediction. Additionally, a Slope-Aware Loss (SAL) is proposed to dynamically assign adaptive weights for the regression head, which takes into account the influence of different slopes on the offset of implant position. Extensive experiments on a dental implant dataset via the five-fold cross-validation demonstrated that the proposed TCSloT achieves state-of-the-art performance.

\section*{Acknowledgment}
This work was supported by the National Natural Science Foundation of China under Grant 82261138629; Guangdong Basic and Applied Basic Research Foundation under Grant 2023A1515010688 and 2021A1515220072; Shenzhen Municipal Science and Technology Innovation Council under Grant JCYJ20220531101412030 and JCYJ20220530155811025.

\bibliographystyle{ieee_fullname}
\bibliography{ref}

\end{document}